\documentclass{article}
\PassOptionsToPackage{numbers, compress}{natbib}
\usepackage[final,nonatbib]{nips_2016}
\usepackage[utf8]{inputenc}
\usepackage[T1]{fontenc}
\usepackage{hyperref}
\usepackage{url}
\usepackage{booktabs}
\usepackage{nicefrac}
\usepackage{microtype}
\usepackage{amsmath}
\usepackage{amsthm}
\usepackage{amssymb}
\usepackage{amsfonts}
\usepackage{bm}
\usepackage{graphicx}
\usepackage{subfigure} 
\usepackage{makecell}

\title{Unsupervised Domain Adaptation with Residual Transfer Networks}

\author{
Mingsheng Long$^\dag$, Han Zhu$^\dag$, Jianmin Wang$^\dag$, and Michael I. Jordan$^\sharp$\\
$^\dag$KLiss, MOE; TNList; School of Software, Tsinghua University, China\\
$^\sharp$University of California, Berkeley, Berkeley, USA\\
{\tt \{mingsheng,jimwang\}@tsinghua.edu.cn, zhuhan10@gmail.com, jordan@berkeley.edu}
}

\begin{document} 

\maketitle

\begin{abstract} 
The recent success of deep neural networks relies on massive amounts of labeled data. For a target task where labeled data is unavailable, domain adaptation can transfer a learner from a different source domain. In this paper, we propose a new approach to domain adaptation in deep networks that can jointly learn adaptive classifiers and transferable features from labeled data in the source domain and unlabeled data in the target domain. We relax a shared-classifier assumption made by previous methods and assume that the source classifier and target classifier differ by a residual function. We enable classifier adaptation by plugging several layers into deep network to explicitly learn the residual function with reference to the target classifier. We fuse features of multiple layers with tensor product and embed them into reproducing kernel Hilbert spaces to match distributions for feature adaptation. The adaptation can be achieved in most feed-forward models by extending them with new residual layers and loss functions, which can be trained efficiently via back-propagation. Empirical evidence shows that the new approach outperforms state of the art methods on standard domain adaptation benchmarks.
\end{abstract} 

\section{Introduction}
Deep networks have significantly improved the state of the art for a wide variety of machine-learning problems and applications. Unfortunately, these impressive gains in performance come only when massive amounts of labeled data are available for supervised training. Since manual labeling of sufficient training data for diverse application domains on-the-fly is often prohibitive, for problems short of labeled data, there is strong incentive to establishing effective algorithms to reduce the labeling consumption, typically by leveraging off-the-shelf labeled data from a different but related source domain. However, this learning paradigm suffers from the shift in data distributions across different domains, which poses a major obstacle in adapting predictive models for the target task \cite{cite:TKDE10TLSurvey}.

Domain adaptation \cite{cite:TKDE10TLSurvey} is machine learning under the shift between training and test distributions. A rich line of approaches to domain adaptation aim to bridge the source and target domains by learning domain-invariant feature representations without using target labels, so that the classifier learned from the source domain can be applied to the target domain. Recent studies have shown that deep networks can learn more transferable features for domain adaptation \cite{cite:ICML14DeCAF,cite:NIPS14CNN}, by disentangling explanatory factors of variations behind domains. Latest advances have been achieved by embedding domain adaptation in the pipeline of deep feature learning which can extract domain-invariant representations \cite{cite:Arxiv14DDC,cite:ICML15DAN,cite:ICML15RevGrad,cite:ICCV15SDT}.

The previous deep domain adaptation methods work under the assumption that the source classifier can be directly transferred to the target domain upon the learned domain-invariant feature representations. This assumption is rather strong as in practical applications, it is often infeasible to check whether the source classifier and target classifier can be shared or not. Hence we focus in this paper on a more general, and safe, domain adaptation scenario in which the source classifier and target classifier differ by a small perturbation function. The goal of this paper is to simultaneously learn adaptive classifiers and transferable features from labeled data in the source domain and unlabeled data in the target domain by embedding the adaptations of both classifiers and features in a unified deep architecture.

Motivated by the state of the art deep residual learning \cite{cite:CVPR16DRL}, winner of the ImageNet ILSVRC 2015 challenge, we propose a new Residual Transfer Network (RTN) approach to domain adaptation in deep networks which can simultaneously learn adaptive classifiers and transferable features. We relax the shared-classifier assumption made by previous methods and assume that the source and target classifiers differ by a small residual function. We enable classifier adaptation by plugging several layers into deep networks to explicitly learn the residual function with reference to the target classifier. In this way, the source classifier and target classifier can be bridged tightly in the back-propagation procedure. The target classifier is tailored to the target data better by exploiting the low-density separation criterion. 
We fuse features of multiple layers with tensor product and embed them into reproducing kernel Hilbert spaces to match distributions for feature adaptation. 
The adaptation can be achieved in most feed-forward models by extending them with new residual layers and loss functions, and can be trained efficiently using standard back-propagation. Extensive evidence suggests that the RTN approach outperforms several state of art methods on standard domain adaptation benchmarks.

\section{Related Work}
Domain adaptation \cite{cite:TKDE10TLSurvey} builds models that can bridge different domains or tasks, which mitigates the burden of manual labeling for machine learning~\cite{cite:TNN11TCA,cite:TPAMI12DTMKL,cite:ICML13TCS,cite:NIPS14FTL}, computer vision \cite{cite:ECCV10Office,cite:CVPR12GFK,cite:NIPS14LSDA} and natural language processing \cite{cite:JMLR11MTLNLP}. The main technical problem of domain adaptation is that the domain discrepancy in probability distributions of different domains should be formally reduced. Deep neural networks can learn abstract representations that disentangle different explanatory factors of variations behind data samples \cite{cite:TPAMI13DLSurvey} and manifest invariant factors underlying different populations that transfer well from original tasks to similar novel tasks \cite{cite:NIPS14CNN}. Thus deep neural networks have been explored for domain adaptation \cite{cite:ICML11DADL,cite:CVPR13MidLevel,cite:NIPS14LSDA}, multimodal and multi-task learning \cite{cite:JMLR11MTLNLP,cite:ICML11MDL}, where significant performance gains have been witnessed relative to prior shallow transfer learning methods. 

However, recent advances show that deep networks can learn abstract feature representations that can only reduce, but not remove, the cross-domain discrepancy \cite{cite:ICML11DADL,cite:Arxiv14DDC}. Dataset shift has posed a bottleneck to the transferability of deep features, resulting in statistically unbounded risk for target tasks \cite{cite:COLT09DAT,cite:ML10DAT}. Some recent work addresses the aforementioned problem by deep domain adaptation, which bridges the two worlds of deep learning and domain adaptation \cite{cite:Arxiv14DDC,cite:ICML15DAN,cite:ICML15RevGrad,cite:ICCV15SDT}. They extend deep convolutional networks (CNNs) to domain adaptation either by adding one or multiple adaptation layers through which the mean embeddings of distributions are matched \cite{cite:Arxiv14DDC,cite:ICML15DAN}, or by adding a fully connected subnetwork as a domain discriminator whilst the deep features are learned to confuse the domain discriminator in a domain-adversarial training paradigm \cite{cite:ICML15RevGrad,cite:ICCV15SDT}. While performance was significantly improved, these state of the art methods may be restricted by the  assumption that under the learned domain-invariant feature representations, the source classifier can be directly transferred to the target domain. In particular, this assumption may not hold when the source classifier and target classifier cannot be shared. As theoretically studied in \cite{cite:ML10DAT}, when the combined error of the ideal joint hypothesis is large, then there is no single classifier that performs well on both source and target domains, so we cannot find a good target classifier by directly transferring from the source domain.

This work is primarily motivated by He et al.~\cite{cite:CVPR16DRL}, the winner of the ImageNet ILSVRC 2015 challenge. They present deep residual learning to ease the training of very deep networks (hundreds of layers), termed \emph{residual nets}. The residual nets explicitly reformulate the layers as learning residual functions $\Delta F(\mathbf{x})$ with reference to the layer inputs $\mathbf{x}$, instead of directly learning the unreferenced  functions $F(\mathbf{x}) = \Delta F(\mathbf{x}) + \mathbf{x}$. The method focuses on standard deep learning in which training data and test data are drawn from identical distributions, hence it cannot be directly applied to domain adaptation. In this paper, we propose to bridge the source classifier $f_S(\mathbf{x})$ and target classifier $f_T(\mathbf{x})$ by the residual layers such that the classifier mismatch across domains can be explicitly modeled by the residual functions $\Delta F(\mathbf{x})$ in a deep learning architecture. Although the idea of adapting source classifier to target domain by adding a perturbation function has been studied by \cite{cite:MM07ASVM,cite:ICML09DAM,cite:TPAMI12AMKL}, these methods require target labeled data to learn the perturbation function, which cannot be applied to unsupervised domain adaptation, the focus of this study. Another distinction is that their perturbation function is defined in the input space $\mathbf{x}$, while the input to our residual function is the target classifier $f_T(\mathbf{x})$, which can capture the connection between the source and target classifiers more effectively.

\section{Residual Transfer Networks}
In unsupervised domain adaptation problem, we are given a source domain $\mathcal{D}_s = \{(\mathbf{x}_i^s,y^s_i)\}_{i=1}^{n_s}$ of $n_s$ labeled examples and a target domain ${{\cal D}_t} = \{ {\bf{x}}_j^t\} _{j = 1}^{{n_t}}$ of $n_t$ unlabeled examples. The source domain and target domain are sampled from different probability distributions $p$ and $q$ respectively, and $p \ne q$. The goal of this paper is to design a deep neural network that enables learning of transfer classifiers $y = f_s\left( {\bf{x}} \right)$ and $y = f_t\left( {\bf{x}} \right)$ to close the source-target discrepancy, such that the expected target risk ${R_t}\left( f_t \right) = {\mathbb{E} _{\left( {{\mathbf{x}},y} \right) \sim q}}\left[ {f_t \left( {\mathbf{x}} \right) \ne y} \right]$ can be bounded by leveraging the source domain supervised data.

The challenge of unsupervised domain adaptation arises in that the target domain has no labeled data, while the source classifier $f_s$ trained on source domain $\mathcal{D}_s$ cannot be directly applied to the target domain $\mathcal{D}_t$ due to the distribution discrepancy $p(\mathbf{x},y) \ne q(\mathbf{x},y)$. The distribution discrepancy may give rise to mismatches in both features and classifiers, i.e. $p(\mathbf{x}) \ne q(\mathbf{x})$ and $f_s(\mathbf{x}) \ne f_t(\mathbf{x})$. Both mismatches should be fixed by joint adaptation of features and classifiers to enable effective domain adaptation. Classifier adaptation is more difficult than feature adaptation because it is directly related to the labels but the target domain is fully unlabeled. Note that the state of the art deep feature adaptation methods \cite{cite:ICML15DAN,cite:ICML15RevGrad,cite:ICCV15SDT} generally assume classifiers can be shared on adapted deep features. This paper assumes $f_s \ne f_t$ and presents an end-to-end deep learning framework for classifier adaptation.

Deep networks \cite{cite:TPAMI13DLSurvey} can learn distributed, compositional, and abstract representations for natural data such as image and text. This paper addresses unsupervised domain adaptation within deep networks for jointly learning transferable features and adaptive classifiers. We extend deep convolutional networks (CNNs), i.e. AlexNet \cite{cite:NIPS12CNN}, to novel residual transfer networks (RTNs) as shown in Figure~\ref{fig:RTN}. Denote by $f_s({\bf x})$ the source classifier, and the empirical error of CNN on source domain data $\mathcal{D}_s$ is
\begin{equation}\label{eqn:CNN}
	\mathop {\min }\limits_{f_s} \frac{1}{{{n_s}}}\sum\limits_{i = 1}^{{n_s}} {L\left( {f_s\left( {{\bf{x}}_i^s} \right),{\bf y}_i^s} \right)} ,
\end{equation}
where $L(\cdot,\cdot)$ is the cross-entropy loss function. Based on the quantification study of feature transferability in deep convolutional networks \cite{cite:NIPS14CNN}, convolutional layers can learn generic features that are transferable across domains \cite{cite:NIPS14CNN}. Hence we opt to fine-tune, instead of directly adapt, the features of convolutional layers when transferring pre-trained deep models from source domain to target domain.

\subsection{Feature Adaptation}
Deep features learned by CNNs can disentangle explanatory factors of variations behind data distributions to boost knowledge transfer \cite{cite:CVPR13MidLevel,cite:TPAMI13DLSurvey}. However, the latest literature findings reveal that deep features can reduce, but not remove, the cross-domain distribution discrepancy \cite{cite:NIPS14CNN}, which motivates the state of the art deep feature adaptation methods \cite{cite:ICML15DAN,cite:ICML15RevGrad,cite:ICCV15SDT}. Deep features in standard CNNs must eventually transition from general to specific along the network, and the transferability of features and classifiers will decrease when the cross-domain discrepancy increases \cite{cite:NIPS14CNN}. In other words, the shifts in the data distributions linger even after multilayer feature abstractions.
In this paper, we perform feature adaptation by matching the feature distributions of multiple layers $\ell \in {\cal L}$ across domains. We reduce feature dimensions by adding a bottleneck layer $fcb$ on top of the last feature layer of CNNs, and then fine-tune CNNs on source labeled examples such that the feature distributions of the source and target are made similar under new feature representations in multiple layers $\mathcal{L} = \{fcb, fcc\}$, as shown in Figure~\ref{fig:RTN}. To adapt multiple feature layers effectively, we propose the tensor product between features of multiple layers to perform lossless multi-layer feature fusion, i.e. ${\mathbf{z}}_i^s \triangleq { \otimes _{\ell  \in \mathcal{L}}}{\mathbf{x}}_i^{s\ell }$ and ${\mathbf{z}}_j^t \triangleq { \otimes _{\ell  \in \mathcal{L}}}{\mathbf{x}}_j^{t\ell}$. We then perform feature adaptation by minimizing the Maximum Mean Discrepancy (MMD) \cite{cite:JMLR12MMD} between source and target domains using the fusion features (dubbed tensor MMD) as
\begin{equation}\label{eqn:MMD}
  \mathop {\min }\limits_{{f_s},{f_t}} {D_\mathcal{L}}\left( {{\mathcal{D}_s},{\mathcal{D}_t}} \right) = \sum\limits_{i = 1}^{{n_s}} {\sum\limits_{j = 1}^{{n_s}} {\frac{{k\left( {{\mathbf{z}}_i^s,{\mathbf{z}}_j^s} \right)}}{{n_s^2}}} }  + \sum\limits_{i = 1}^{{n_t}} {\sum\limits_{j = 1}^{{n_t}} {\frac{{k\left( {{\mathbf{z}}_i^t,{\mathbf{z}}_j^t} \right)}}{{n_t^2}}} }  - 2\sum\limits_{i = 1}^{{n_s}} {\sum\limits_{j = 1}^{{n_t}} {\frac{{k\left( {{\mathbf{z}}_i^s,{\mathbf{z}}_j^t} \right)}}{{{n_s}{n_t}}}} },
\end{equation}
where the characteristic kernel $k( {{{\mathbf{z}}},{{\mathbf{z}'}}} ) = e^{{ - {{\left\| {\rm{vec}( {\mathbf{z}} ) - \rm{vec}( {{\mathbf{z'}}} )} \right\|}^2}/b }}$ is the Gaussian kernel function defined on the vectorization of tensors ${\mathbf{z}}$ and $\mathbf{z}'$ with bandwidth parameter $b$. Different from DAN \cite{cite:ICML15DAN} that adapts multiple feature layers using multiple MMD penalties, this paper adapts multiple feature layers by first fusing them and then adapting the fused features. The advantage of our method against DAN \cite{cite:ICML15DAN} is that our method can capture full interactions across multilayer features and facilitate easier model selection, while DAN \cite{cite:ICML15DAN} needs $\left| \mathcal{L} \right|$ independent MMD penalties for adapting $\left| \mathcal{L} \right|$ layers.

\subsection{Classifier Adaptation}
As feature adaptation cannot remove the mismatch in classification models, we further perform classifier adaptation to learn transfer classifiers that make domain adaptation more effective. Although the source classifier $f_s(\mathbf{x})$ and target classifier $f_t(\mathbf{x})$ are different, $f_s(\mathbf{x}) \ne f_t(\mathbf{x})$, they should be related to ensure the feasibility of domain adaptation. It is reasonable to assume that $f_s(\mathbf{x})$ and $f_t(\mathbf{x})$ differ only by a small perturbation function $\Delta f(\mathbf{x})$. Prior work on classifier adaptation \cite{cite:MM07ASVM,cite:ICML09DAM,cite:TPAMI12AMKL} assumes that $f_t(\mathbf{x}) = f_s(\mathbf{x}) + \Delta f(\mathbf{x})$, where the perturbation $\Delta f(\mathbf{x})$ is a function of input feature $\mathbf{x}$. However, these methods require target labeled data to learn the perturbation function, which cannot be applied to unsupervised domain adaptation where target domain has no labeled data. How to bridge $f_s(\mathbf{x})$ and $f_t(\mathbf{x})$ in a framework is a key challenge of unsupervised domain adaptation. We postulate that the perturbation function $\Delta f(\mathbf{x})$ can be learned jointly from the source labeled data and target unlabeled data, given that the source classifier and target classifier are properly connected.

\begin{figure*}[tbp]
  \centering
  \includegraphics[width=1.0\textwidth]{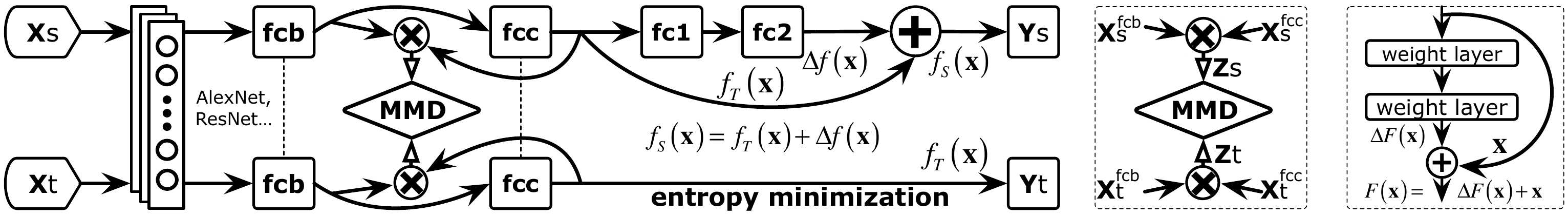}
  \caption{(left) Residual Transfer Network (RTN) for domain adaptation, based on well-established architectures. Due to dataset shift, (1) the last-layer features are tailored to domain-specific structures that are not safely transferable, hence we add a bottleneck layer $fcb$ that is adapted jointly with the classifier layer $fcc$ by the tensor MMD module; (2) Supervised classifiers are not safely transferable, hence we bridge them by the residual layers $fc1$--$fc2$ so that ${f_S}\left( {\mathbf{x}} \right) = {f_T}\left( {\mathbf{x}} \right) + \Delta f\left( \mathbf{x} \right)$. (middle) The tensor MMD module for multi-layer feature adaptation. (right) The building block for deep residual learning; Instead of using the residual block to model feature mappings, we use it to bridge the source classifier $f_S(\mathbf{x})$ and target classifier $f_T(\mathbf{x})$ with $\mathbf{x} \triangleq f_T(\mathbf{x})$, ${F}(\mathbf{x}) \triangleq f_S(\mathbf{x})$, and $\Delta {F}(\mathbf{x}) \triangleq \Delta f(\mathbf{x})$.}
  \label{fig:RTN}
  \vspace{-10pt}
\end{figure*}

To enable classifier adaptation, consider fitting $F(\mathbf{x})$ as an original mapping by a few stacked layers (convolutional or fully connected layers) in Figure~\ref{fig:RTN} (right), where $\mathbf{x}$ denotes the inputs to the first of these layers \cite{cite:CVPR16DRL}. If one hypothesizes that multiple nonlinear layers can asymptotically approximate complicated functions, then it is equivalent to hypothesize that they can asymptotically approximate the residual functions, i.e. $F(\mathbf{x}) - \mathbf{x}$. Rather than expecting stacked layers to approximate $F(\mathbf{x})$, one explicitly lets these layers approximate a residual function $\Delta F(\mathbf{x}) \triangleq F(\mathbf{x}) - \mathbf{x}$, with the original function being $\Delta F(\mathbf{x}) + \mathbf{x}$. The operation $\Delta F(\mathbf{x}) + \mathbf{x}$ is performed by a \emph{shortcut} connection and an element-wise addition, while the residual function is parameterized by residual layers within each residual block. Although both forms are able to asymptotically approximate the desired functions, the ease of learning is different. In reality, it is unlikely that identity mappings are optimal, but it should be easier to find the perturbations with reference to an identity mapping, than to learn the function as new. The residual learning is the key to the successful training of very deep networks. The deep residual network (ResNet) framework \cite{cite:CVPR16DRL} bridges the inputs and outputs of the residual layers by the shortcut connection (identity mapping) such that $F(\mathbf{x}) = \Delta F(\mathbf{x}) + \mathbf{x}$, which eases the learning of residual function $\Delta F(\mathbf{x})$ (similar to the perturbation function across the source and target classifiers). 

Based on this key observation, we extend the CNN architectures (Figure~\ref{fig:RTN}, left) by plugging in the residual block (Figure~\ref{fig:RTN}, right). We reformulate the residual block to bridge the source classifier $f_S(\mathbf{x})$ and target classifier $f_T(\mathbf{x})$ by letting $\mathbf{x} \triangleq f_T(\mathbf{x})$, $F(\mathbf{x}) \triangleq f_S(\mathbf{x})$, and $\Delta F(\mathbf{x}) \triangleq \Delta f(\mathbf{x})$. Note that $f_S(\mathbf{x})$ is the outputs of the element-wise addition operator and $f_T(\mathbf{x})$ is the outputs of the target-classifier layer $fcc$, both before softmax activation $\sigma(\cdot)$, ${f_s}\left( {\mathbf{x}} \right) \triangleq \sigma \left( {{f_S}\left( {\mathbf{x}} \right)} \right),{f_t} \left( {\mathbf{x}} \right) \triangleq \sigma \left( {{f_T}\left( {\mathbf{x}} \right)} \right)$. We can connect the source classifier and target classifier (before activation) by the residual block as
\begin{equation}\label{eqn:Residual}
	{f_S}\left( {\mathbf{x}} \right) = {f_T}\left( {\mathbf{x}} \right) + \Delta f\left( \mathbf{x} \right),
\end{equation}
where we use functions $f_S$ and $f_T$ before softmax for residual block to ensure that the final classifiers $f_s$ and $f_t$ will output probabilities. Residual layers $fc1$--$fc2$ are fully-connected layers with $c \times c$ units, where $c$ is the number of classes. We set the source classifier $f_S$ as the outputs of the residual block to make it better trainable from the source-labeled data by deep residual learning \cite{cite:CVPR16DRL}. In other words, if we set $f_T$ as the outputs of the residual block, then we may be unable to learn it successfully as we do not have target labeled data and thus standard back-propagation will not work. Deep residual learning \cite{cite:CVPR16DRL} ensures to output valid classifiers $\left| \Delta f\left( \mathbf{x} \right) \right| \ll \left| {{f_T}\left( {\mathbf{x}} \right)} \right| \approx \left| {{f_S}\left( {\mathbf{x}} \right)} \right|$, and more importantly, makes the perturbation function $\Delta f\left( \mathbf{x} \right)$ dependent on both the target classifier $f_T(\mathbf{x})$ (due to the functional dependency) as well as the source classifier $f_S(\mathbf{x})$ (due to the back-propagation pipeline).

Although we successfully cast the classifier adaptation into the residual learning framework while the residual learning framework tends to make the target classifier $f_t(\mathbf{x})$ not deviate much from the source classifier $f_s(\mathbf{x})$, we still cannot guarantee that $f_t(\mathbf{x})$ will fit the target-specific structures well. To address this problem, we further exploit the entropy minimization principle \cite{cite:NIPS04SSLEM} for refining the classifier adaptation, which encourages the low-density separation between classes by minimizing the entropy of class-conditional distribution $f^t_j(\mathbf{x}^t_i) = p(y^t_i=j|\mathbf{x}^t_i;f_t)$ on target domain data $\mathcal{D}_t$ as
\begin{equation}\label{eqn:Entropy}
	\mathop {\min }\limits_{f_t} \frac{1}{{{n_t}}}\sum\limits_{i = 1}^{{n_t}} {H\left( {f_t\left( {{\bf{x}}_i^t} \right)} \right)} ,
\end{equation}
where $H(\cdot)$ is the entropy function of class-conditional distribution ${f_t\left( {{\bf{x}}_i^t} \right)}$ defined as $ H\left( {f_t\left( {{\bf{x}}_i^t} \right)} \right) =  - \sum\nolimits_{j = 1}^c {{f^t_j}\left( {{\bf{x}}_i^t} \right)\log {f^t_j}\left( {{\bf{x}}_i^t} \right)} $, $c$ is the number of classes, and ${f^t_j}\left( {{\mathbf{x}}_i^t} \right)$ is the probability of predicting point $\mathbf{x}_i^t$ to class $j$. By minimizing entropy penalty~\eqref{eqn:Entropy}, the target classifier $f_t(\mathbf{x})$ is made directly accessible to target-unlabeled data and will amend itself to pass through the target low-density regions.

\subsection{Residual Transfer Network}
To enable effective unsupervised domain adaptation, we propose  Residual Transfer Network (RTN), which jointly learns transferable features and adaptive classifiers by integrating deep feature learning \eqref{eqn:CNN}, feature adaptation \eqref{eqn:MMD}, and classifier adaptation \eqref{eqn:Residual}--\eqref{eqn:Entropy} in an end-to-end deep learning framework,
\begin{equation}\label{eqn:model}
\begin{aligned}
	\mathop {\min }\limits_{{f_S} = {f_T} + \Delta f} \frac{1}{{{n_s}}} & \sum\limits_{i = 1}^{{n_s}} {L\left( {{f_s}\left( {{\mathbf{x}}_i^s} \right),y_i^s} \right)} \\
  + \ \frac{\gamma }{{{n_t}}} & \sum\limits_{i = 1}^{{n_t}} {H\left( {{f_t}\left( {{\mathbf{x}}_i^t} \right)} \right)}  \\
	+ \ \lambda & \ {D_{\cal L}\left( {\mathcal{D}_s ,\mathcal{D}_t } \right)},
\end{aligned}
\end{equation}
where $\lambda$ and $\gamma$ are the tradeoff parameters for the tensor MMD penalty~\eqref{eqn:MMD} and entropy penalty~\eqref{eqn:Entropy} respectively. The proposed RTN model \eqref{eqn:model} is enabled to learn both transferable features and  adaptive classifiers. As classifier adaptation proposed in this paper and feature adaptation studied in \cite{cite:ICML15DAN,cite:ICML15RevGrad} are tailored to adapt different layers of deep networks, they can complement each other to establish better performance. Since training deep CNNs requires a large amount of labeled data that is prohibitive for many domain adaptation applications, we start with the CNN models pre-trained on ImageNet 2012 data and fine-tune it as \cite{cite:ICML15DAN}. The training of RTN mainly follows standard back-propagation, with the residual transfer layers for classifier adaptation as \cite{cite:CVPR16DRL}. Note that, the optimization of tensor MMD penalty \eqref{eqn:MMD} requires carefully-designed algorithm to establish linear-time training, as detailed in \cite{cite:ICML15DAN}. We also adopt bilinear pooling \cite{cite:CVPR15BCNN} to reduce the dimensions of fusion features in tensor MMD \eqref{eqn:MMD}.

\section{Experiments}
We evaluate the residual transfer network against state of the art transfer learning and deep learning methods. Codes and datasets will be available at \url{https://github.com/thuml/transfer-caffe}.

\subsection{Setup}
\textbf{Office-31} \cite{cite:ECCV10Office} is a benchmark for domain adaptation, comprising 4,110 images in 31 classes collected from three distinct domains: \textit{Amazon} (\textbf{A}), which contains images downloaded from \url{amazon.com}, \textit{Webcam} (\textbf{W}) and \textit{DSLR} (\textbf{D}), which contain images taken by web camera and digital SLR camera with different photographical settings, respectively. To enable unbiased evaluation, we evaluate all methods on all six transfer tasks \textbf{A} $\rightarrow$ \textbf{W}, \textbf{D} $\rightarrow$ \textbf{W}, \textbf{W} $\rightarrow$ \textbf{D}, \textbf{A} $\rightarrow$ \textbf{D}, \textbf{D} $\rightarrow$ \textbf{A} and \textbf{W} $\rightarrow$ \textbf{A} as in \cite{cite:ICML15DAN,cite:ICCV15SDT}.

\textbf{Office-Caltech} \cite{cite:CVPR12GFK} is built by selecting the 10 common categories shared by \textit{Office-31} and \textit{Caltech-256} (\textbf{C}), and is widely used by previous methods \cite{cite:CVPR12GFK,cite:AAAI16CORAL}. We can build 12 transfer tasks: \textbf{A} $\rightarrow$ \textbf{W}, \textbf{D} $\rightarrow$ \textbf{W}, \textbf{W} $\rightarrow$ \textbf{D}, \textbf{A} $\rightarrow$ \textbf{D}, \textbf{D} $\rightarrow$ \textbf{A}, \textbf{W} $\rightarrow$ \textbf{A}, \textbf{A} $\rightarrow$ \textbf{C}, \textbf{W} $\rightarrow$ \textbf{C}, \textbf{D} $\rightarrow$ \textbf{C}, \textbf{C} $\rightarrow$ \textbf{A}, \textbf{C} $\rightarrow$ \textbf{W}, and \textbf{C} $\rightarrow$ \textbf{D}. While \emph{Office-31} has more categories and is more difficult for domain adaptation algorithms, \emph{Office-Caltech} provides more transfer tasks to enable an unbiased look at dataset bias \cite{cite:CVPR11DB}. We adopt DeCAF$_7$ \cite{cite:ICML14DeCAF} features for shallow transfer methods and original images for deep adaptation methods.

We compare with both conventional and the state of the art transfer learning and deep learning methods: Transfer Component Analysis (\textbf{TCA}) \cite{cite:TNN11TCA}, Geodesic Flow Kernel (\textbf{GFK}) \cite{cite:CVPR12GFK}, Deep Convolutional Neural Network (\textbf{AlexNet} \cite{cite:NIPS12CNN}), Deep Domain Confusion (\textbf{DDC}) \cite{cite:Arxiv14DDC}, Deep Adaptation Network (\textbf{DAN}) \cite{cite:ICML15DAN}, and Reverse Gradient (\textbf{RevGrad}) \cite{cite:ICML15RevGrad}. TCA is a conventional transfer learning method based on MMD-regularized Kernel PCA. GFK is a manifold learning method that interpolates across an infinite number of intermediate subspaces to bridge domains. DDC is the first method that maximizes domain invariance by adding to AlexNet an adaptation layer using linear-kernel MMD \cite{cite:JMLR12MMD}. DAN learns more transferable features by embedding deep features of multiple task-specific layers to reproducing kernel Hilbert spaces (RKHSs) and matching different distributions optimally using multi-kernel MMD. RevGrad improves domain adaptation by making the source and target domains indistinguishable for a discriminative domain classifier via an adversarial training paradigm. 

To go deeper with the efficacy of classifier adaptation (residual transfer block) and feature adaptation (tensor MMD module), we perform ablation study by evaluating several variants of RTN: (1) \textbf{RTN (mmd)}, which adds the tensor MMD module to AlexNet; (2) \textbf{RTN (mmd+ent)}, which further adds the entropy penalty to AlexNet; (3) \textbf{RTN (mmd+ent+res)}, which further adds the residual module to AlexNet. Note that RTN (mmd) improves DAN \cite{cite:ICML15DAN} by replacing the multiple MMD penalties in DAN by a single tensor MMD penalty in RTN (mmd), which facilitates much easier parameter selection.

We follow standard protocols and use all labeled source data and all unlabeled target data for domain adaptation \cite{cite:ICML15DAN}. We compare average classification accuracy of each transfer task using three random experiments. For MMD-based methods (TCA, DDC, DAN, and RTN), we use Gaussian kernel with bandwidth $b$ set to median pairwise squared distances on training data, i.e. median heuristic \cite{cite:JMLR12MMD}. As there are no target labeled data in unsupervised domain adaptation, model selection proves difficult. For all methods, we perform cross-valuation on labeled source data to select candidate parameters, then conduct validation on transfer task $\textbf{A} \rightarrow \textbf{W}$ by requiring one labeled example per category from target domain \textbf{W} as the validation set, and fix the selected parameters throughout all transfer tasks.

We implement all deep methods based on the \textbf{Caffe} deep-learning framework, and fine-tune from Caffe-provided models of AlexNet \cite{cite:NIPS12CNN} pre-trained on ImageNet. For RTN, We fine-tune all the feature layers, train bottleneck layer $fcb$, classifier layer $fcc$ and residual layers $fc1$--$fc2$, all through standard back-propagation. Since these new layers are trained from scratch, we set their learning rate to be 10 times that of the other layers. We use mini-batch stochastic gradient descent (SGD) with momentum of 0.9 and the learning rate annealing strategy implemented in RevGrad \cite{cite:ICML15RevGrad}: the learning rate is not selected through a grid search due to high computational cost---it is adjusted during SGD using the following formula: ${\eta _p} = \frac{{{\eta _0}}}{{{{\left( {1 + \alpha p} \right)}^\beta }}}$, where $p$ is the training progress linearly changing from $0$ to $1$, $\eta_0 = 0.01, \alpha=10$ and $\beta=0.75$, which is optimized for low error on the source domain. As RTN can work stably across different transfer tasks, the MMD penalty parameter $\lambda$ and entropy penalty $\gamma$ are first selected on $\textbf{A} \rightarrow \textbf{W}$ and then fixed as $\lambda=0.3, \gamma=0.3$ for all other transfer tasks.

\subsection{Results}
The classification accuracy results on the six transfer tasks of \textit{Office-31} are shown in Table \ref{table:office31}, and the results on the twelve transfer tasks of \textit{Office-Caltech} are shown in Table \ref{table:office-caltech}. The RTN model based on AlexNet (Figure~\ref{fig:RTN}) outperforms all comparison methods on most transfer tasks. In particular, RTN substantially improves the accuracy on hard transfer tasks, e.g. \textbf{A $\rightarrow$ W} and \textbf{C $\rightarrow$ W}, where the source and target domains are very different, and achieves comparable accuracy on easy transfer tasks, \textbf{D $\rightarrow$ W} and \textbf{W $\rightarrow$ D}, where source and target domains are similar \cite{cite:ECCV10Office}. These results suggest that RTN is able to learn more adaptive classifiers and transferable features for safer domain adaptation.

\begin{table}[htp]
    \addtolength{\tabcolsep}{0pt}
    \centering
    \caption{Accuracy on \emph{Office-31} dataset using standard protocol \cite{cite:ICML15DAN} for unsupervised adaptation.}
    \label{table:office31}
    \begin{small}
    \begin{tabular}{cccccccc}
        \Xhline{1pt}
        Method & A $\rightarrow$ W & D $\rightarrow$ W & W $\rightarrow$ D & A $\rightarrow$ D & D $\rightarrow$ A & W $\rightarrow$ A & Avg\\
        \hline
		TCA \cite{cite:TNN11TCA} & 59.0$\pm$0.0 & 90.2$\pm$0.0 & 88.2$\pm$0.0 & 57.8$\pm$0.0 & 51.6$\pm$0.0 & 47.9$\pm$0.0 & 65.8\\
		GFK \cite{cite:CVPR12GFK} & 58.4$\pm$0.0 & 93.6$\pm$0.0 & 91.0$\pm$0.0 & 58.6$\pm$0.0 & \textbf{52.4}$\pm$0.0 & 46.1$\pm$0.0 & 66.7\\
		AlexNet \cite{cite:NIPS12CNN} & 60.6$\pm$0.4 & 95.4$\pm$0.2 & 99.0$\pm$0.1 & 64.2$\pm$0.3 & 45.5$\pm$0.5 & 48.3$\pm$0.5 & 68.8\\
		DDC \cite{cite:Arxiv14DDC} & 61.0$\pm$0.5 & 95.0$\pm$0.3 & 98.5$\pm$0.3 & 64.9$\pm$0.4 & 47.2$\pm$0.5 & 49.4$\pm$0.6 & 69.3\\
		DAN \cite{cite:ICML15DAN} & 68.5$\pm$0.3 & 96.0$\pm$0.1 & 99.0$\pm$0.1 & 66.8$\pm$0.2 & 50.0$\pm$0.4 & 49.8$\pm$0.3 & 71.7\\
		RevGrad \cite{cite:ICML15RevGrad} & 73.0$\pm$0.6 & 96.4$\pm$0.4 & 99.2$\pm$0.3 & - & - & - & -\\
        \hline
        RTN (mmd) & 70.0$\pm$0.4 & 96.1$\pm$0.3 & {99.2}$\pm$0.3 & 67.6$\pm$0.4 & 49.8$\pm$0.4 & 50.0$\pm$0.3 & 72.1 \\				
        RTN (mmd+ent) & 71.2$\pm$0.3 & 96.4$\pm$0.2 & {99.2}$\pm$0.1 & 69.8$\pm$0.2 & 50.2$\pm$0.3 & 50.7$\pm$0.2 & 72.9 \\
        RTN (mmd+ent+res) & \textbf{73.3}$\pm$0.3 & \textbf{96.8}$\pm$0.2 & \textbf{99.6}$\pm$0.1 & \textbf{71.0}$\pm$0.2 & {50.5}$\pm$0.3 & \textbf{51.0}$\pm$0.1 & \textbf{73.7} \\
        \Xhline{1pt}
    \end{tabular}
    \end{small}
\end{table}

\begin{table}[htp]
    \addtolength{\tabcolsep}{-5pt}
    \centering
    \caption{Accuracy on \emph{Office-Caltech} dataset using standard protocol \cite{cite:ICML15DAN} for unsupervised adaptation.}
    \label{table:office-caltech}
    \begin{small}
    \begin{tabular}{cccccccccccccc}
        \Xhline{1pt}
        Method & A$\rightarrow$W & D$\rightarrow$W & W$\rightarrow$D & A$\rightarrow$D & D$\rightarrow$A & W$\rightarrow$A & A$\rightarrow$C & W$\rightarrow$C & D$\rightarrow$C & C$\rightarrow$A & C$\rightarrow$W & C$\rightarrow$D & Avg \\
        \hline
		TCA \cite{cite:TNN11TCA} & 84.4 & 96.9 & 99.4 & 82.8 & 90.4 & 85.6 & 81.2 & 75.5 & 79.6 & 92.1 & 88.1 & 87.9 & 87.0 \\
		GFK \cite{cite:CVPR12GFK} & 89.5 & 97.0 & 98.1 & 86.0 & 89.8 & 88.5 & 76.2 & 77.1 & 77.9 & 90.7 & 78.0 & 77.1 & 85.5 \\
		AlexNet \cite{cite:NIPS12CNN} & 79.5 & 97.7 & \textbf{100.0} & 87.4& 87.1 & 83.8 & 83.0 & 73.0 & 79.0 & 91.9 & 83.7 & 87.1 & 86.1 \\
		DDC \cite{cite:Arxiv14DDC} & 83.1 & 98.1 & \textbf{100.0} & 88.4 & 89.0 & 84.9 & 83.5 & 73.4 & 79.2 & 91.9 & 85.4 & 88.8 & 87.1 \\
		DAN \cite{cite:ICML15DAN} & 91.8 & 98.5 & \textbf{100.0} & 91.7 & 90.0 & 92.1 & 84.1 & 81.2 & 80.3 & 92.0 & 90.6 & 89.3 & 90.1 \\
		\hline
        RTN (mmd) & 93.2 & 98.5 & \textbf{100.0} & 91.7 & 88.0 & 90.7 & 84.0 & 81.3 & 80.4 & 91.0 & 89.8 & 90.4 & 90.0 \\
        RTN (mmd+ent) & 93.8 & 98.6 & \textbf{100.0} & 92.9 & 93.6 & \textbf{92.7} & 87.8 & 84.8 & 83.4 & 93.2 & 96.6 & 93.9 & 92.6 \\
        RTN (mmd+ent+res) & \textbf{95.2} & \textbf{99.2} & \textbf{100.0} & \textbf{95.5} & \textbf{93.8} & 92.5 & \textbf{88.1} & \textbf{86.6} & \textbf{84.6} & \textbf{93.7} & \textbf{96.9} & \textbf{94.2} & \textbf{93.4} \\
        \Xhline{1pt}
    \end{tabular}
    \end{small}
    \vspace{-10pt}
\end{table}

From the results, we can make interesting observations. (1) Standard deep-learning methods (AlexNet) perform comparably with traditional shallow transfer-learning methods with deep DeCAF$_7$ features as input (TCA and GFK). The only difference between these two sets of methods is that AlexNet can take the advantage of supervised fine-tuning on the source-labeled data, while TCA and GFK can take benefits of their domain adaptation procedures. This result confirms the current practice that supervised fine-tuning is important for transferring source classifier to target domain \cite{cite:CVPR13MidLevel}, and sustains the recent discovery that deep neural networks learn abstract feature representation, which can only reduce, but not remove, the cross-domain discrepancy \cite{cite:NIPS14CNN}. This reveals that the two worlds of deep learning and domain adaptation cannot reinforce each other substantially in the two-step pipeline, which motivates carefully-designed deep adaptation architectures to unify them. (2) Deep-transfer learning methods that reduce the domain discrepancy by domain-adaptive deep networks (DDC, DAN and RevGrad) substantially outperform standard deep learning methods (AlexNet) and traditional shallow transfer-learning methods with deep features as the input (TCA and GFK). This confirms that incorporating domain-adaptation modules into deep networks can improve domain adaptation performance. By adapting source-target distributions in multiple task-specific layers using optimal multi-kernel two-sample matching, DAN performs the best in general among the prior deep-transfer learning methods. (3) The proposed residual transfer network (RTN) performs the best and sets up a new state of the art result on these benchmark datasets. Different from all the previous deep-transfer learning methods that only adapt the feature layers of deep neural networks to learn more transferable features, RTN further adapts the classifier layers to bridge the source and target classifiers in an end-to-end residual learning framework, which can correct the classifier mismatch more effectively.

To go deeper into different modules of RTN, we show the results of RTN variants in Tables~\ref{table:office31} and \ref{table:office-caltech}. (1) RTN (mmd) slightly outperforms DAN, but RTN (mmd) has only one MMD penalty parameter while DAN has two or three. Thus the proposed tensor MMD module is effective for adapting multiple feature layers using a single MMD penalty, which is important for easy model selection. (2) RTN (mmd+ent) performs substantially better than RTN (mmd). This highlights the importance of entropy minimization for low-density separation, which exploits the cluster structure of target-unlabeled data such that the target-classifier can be better adapted to the target data. (3) RTN (mmd+ent+res) performs the best across all variants. This highlights the importance of residual transfer of classifier layers for learning more adaptive classifiers. This is critical as in practical applications, there is no guarantee that the source classifier and target classifier can be safely shared. It is worth noting that, the entropy penalty and the residual module should be used together, otherwise the residual function tends to learn useless zero mapping such that the source and target classifiers are nearly identical \cite{cite:CVPR16DRL}.

\subsection{Discussion}
\begin{figure}[tbp]
  \centering
  \subfigure[DAN: \textit{Source}=\textbf{A}]{
    \includegraphics[width=0.18\textwidth]{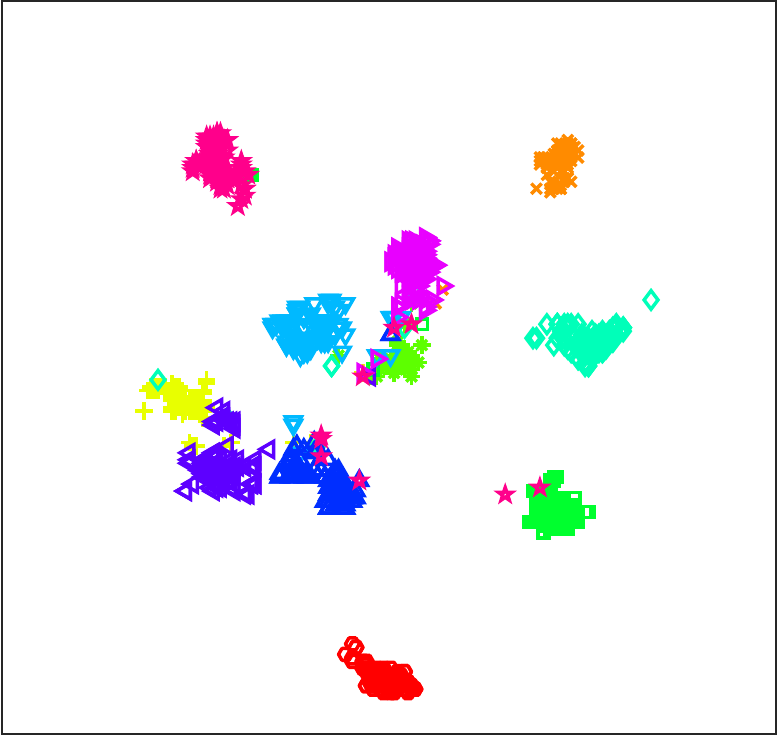}
    \label{fig:dan_s}
  }
  \hfil
  \subfigure[DAN: \textit{Target}=\textbf{W}]{
    \includegraphics[width=0.18\textwidth]{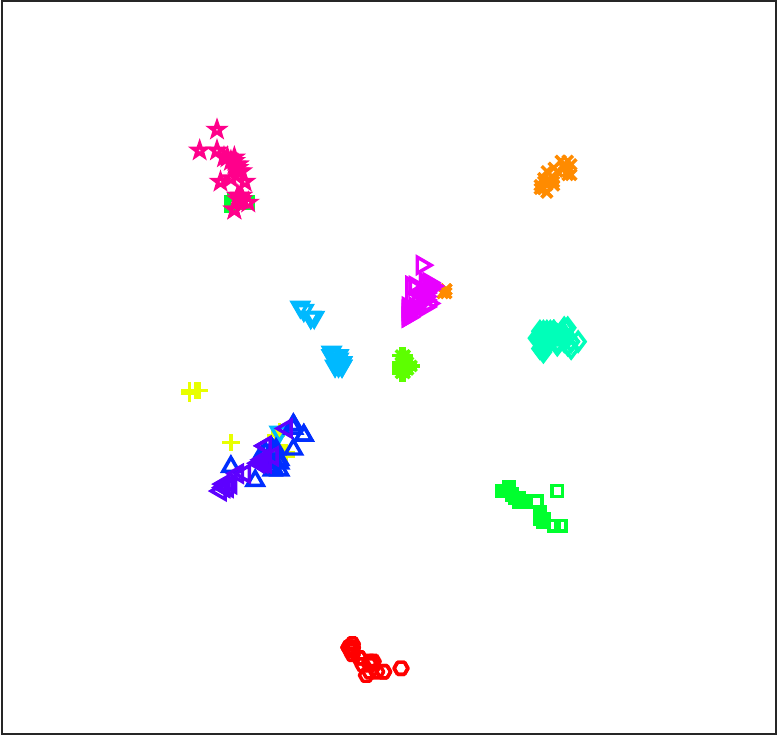}
    \label{fig:dan_t}
  }
  \hfil
  \subfigure[RTN: \textit{Source}=\textbf{A}]{
    \includegraphics[width=0.18\textwidth]{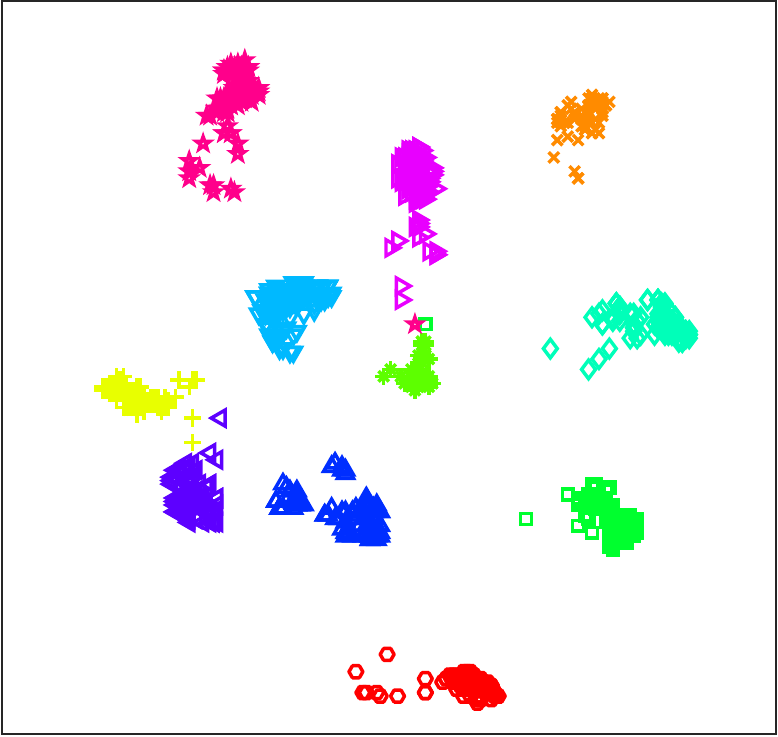}
    \label{fig:rtn_s}
  }
  \hfil
  \subfigure[RTN: \textit{Target}=\textbf{W}]{
    \includegraphics[width=0.18\textwidth]{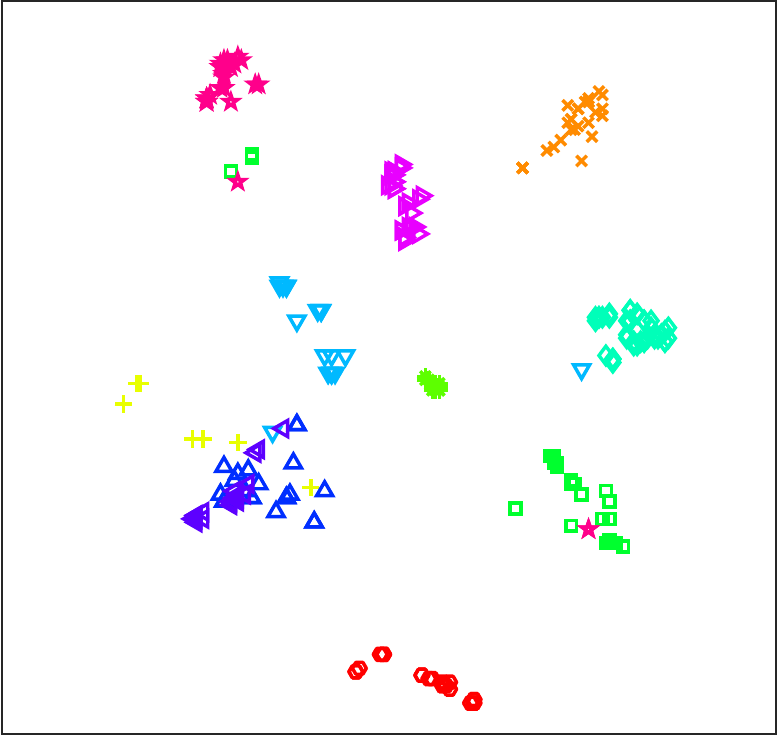}
    \label{fig:rtn_t}
  }
  \caption{Visualization: (a)-(b) t-SNE of DAN predictions; (c)-(d) t-SNE of RTN predictions.}
  \vspace{-10pt}
\end{figure}

\textbf{Predictions Visualization:}
We respectively visualize in Figures~\ref{fig:dan_s}--\ref{fig:rtn_t} the t-SNE embeddings \cite{cite:ICML14DeCAF} of the predictions by DAN and RTN on transfer task \textbf{A} $\rightarrow$ \textbf{W}. 
We can make the following observations. (1) The predictions made by DAN in Figure~\ref{fig:dan_s}--\ref{fig:dan_t} show that the target categories are not well discriminated by the source classifier, which implies that target data is not well compatible with the source classifier. Hence the source and target classifiers should not be assumed to be identical, which has been a common assumption made by all prior deep domain adaptation methods \cite{cite:Arxiv14DDC,cite:ICML15DAN,cite:ICML15RevGrad,cite:ICCV15SDT}. (2) The predictions made by RTN in Figures~\ref{fig:rtn_s}--\ref{fig:rtn_t} show that the target categories are discriminated better by the target classifier (larger class-to-class distances), which suggests that residual transfer of classifiers is a reasonable extension to previous deep feature adaptation methods. RTN simultaneously learns more adaptive classifiers and more transferable features to enable effective domain adaptation.

\begin{figure}[tbp]
  \centering
  \subfigure[Layer Responses]{
    \includegraphics[width=0.26\columnwidth]{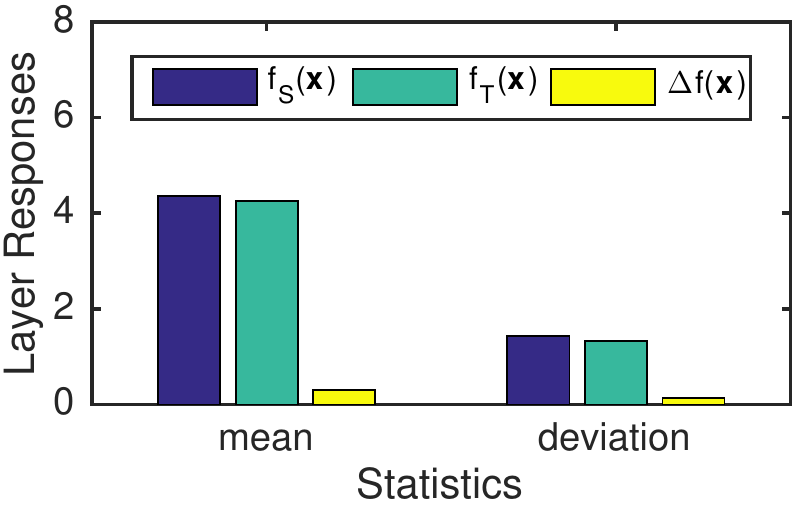}
    \label{fig:response}
  }
  \hfil
  \subfigure[Classifier Shift]{
    \includegraphics[width=0.26\columnwidth]{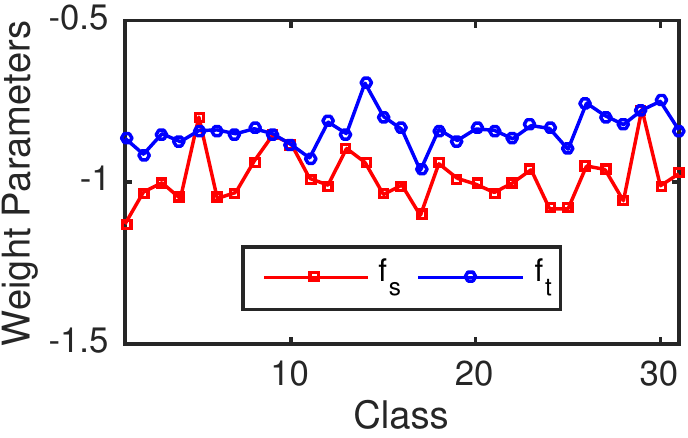}
    \label{fig:mismatch}
  }
  \hfil
  \subfigure[Accuracy w.r.t. $\gamma$]{
    \includegraphics[width=0.26\columnwidth]{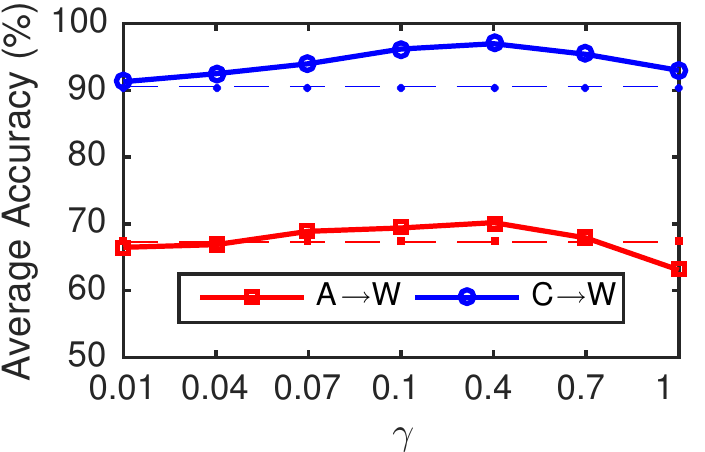}
    \label{fig:sensitivity}
  }
  \caption{(a) layer responses; (b) classifier shift; (c) sensitivity of $\gamma$ (dashed lines show best baselines).}
  \vspace{-10pt}
\end{figure}

\textbf{Layer Responses:}
We show in Figure~\ref{fig:response} the means and standard deviations of the layer responses \cite{cite:CVPR16DRL}, which are the outputs of $f_T(\mathbf{x})$ ($fcc$ layer), $\Delta f(\mathbf{x})$ ($fc2$ layer), and $f_S(\mathbf{x})$ (after element-wise sum operator), respectively. This exposes the response strength of the residual functions. The results show that the residual function $\Delta f(\mathbf{x})$ have generally much smaller responses than the shortcut function $f_T(\mathbf{x})$. These results support our motivation that the residual functions are generally smaller than the non-residual functions, as they characterize the small gap between the source classifier and target classifier. The small residual function can be learned effectively via deep residual learning  \cite{cite:CVPR16DRL}.

\textbf{Classifier Shift:}
To justify that there exists a classifier shift between source classifier $f_s$ and target classifier $f_t$, we train $f_s$ on source domain and $f_t$ on target domain, both provided with labeled data. By taking \textbf{A} as source domain and \textbf{W} as target domain, the weight parameters of the classifiers (e.g. softmax regression) are shown in Figure~\ref{fig:mismatch}, which shows that $f_s$ and $f_t$ are substantially different.

\textbf{Parameter Sensitivity:}
We check the sensitivity of entropy parameter $\gamma$ on transfer tasks \textbf{A} $\rightarrow$ \textbf{W} (31 classes) and \textbf{C} $\rightarrow$ \textbf{W} (10 classes) by varying the parameter in $\{ 0.01, 0.04, 0.07, 0.1, 0.4, 0.7, 1.0 \}$. The results are shown in Figures~\ref{fig:sensitivity}, with the best results of the baselines shown as dashed lines. The accuracy of RTN first increases and then decreases as $\gamma$ varies and demonstrates a desirable bell-shaped curve. This justifies our motivation of jointly learning transferable features and adaptive classifiers by the RTN model, as a good trade-off between them can promote transfer performance.

\section{Conclusion}
This paper presented a novel approach to unsupervised domain adaptation in deep networks, which enables end-to-end learning of adaptive classifiers and transferable features. Similar to many prior domain adaptation techniques, feature adaptation is achieved by matching the distributions of features across domains. However, unlike previous work, the proposed approach also supports classifier adaptation, which is implemented through a new residual transfer module that bridges the source classifier and target classifier. This makes the approach a good complement to existing techniques. The approach can be trained by standard back-propagation, which is scalable and can be implemented by most deep learning package. Future work constitutes semi-supervised domain adaptation extensions.

\section*{Acknowledgments}
This work was supported by the National Natural Science Foundation of China (61502265, 61325008), National Key R\&D Program of China (2016YFB1000701, 2015BAF32B01), and TNList Key Project.

\begin{small}
\bibliography{RTN}

\begin{thebibliography}{10}

\bibitem{cite:TKDE10TLSurvey}
S.~J. Pan and Q.~Yang.
\newblock A survey on transfer learning.
\newblock {\em TKDE}, 22(10):1345--1359, 2010.

\bibitem{cite:ICML14DeCAF}
J.~Donahue, Y.~Jia, O.~Vinyals, J.~Hoffman, N.~Zhang, E.~Tzeng, and T.~Darrell.
\newblock Decaf: A deep convolutional activation feature for generic visual
  recognition.
\newblock In {\em ICML}, 2014.

\bibitem{cite:NIPS14CNN}
J.~Yosinski, J.~Clune, Y.~Bengio, and H.~Lipson.
\newblock How transferable are features in deep neural networks?
\newblock In {\em NIPS}, 2014.

\bibitem{cite:Arxiv14DDC}
E.~Tzeng, J.~Hoffman, N.~Zhang, K.~Saenko, and T.~Darrell.
\newblock Deep domain confusion: Maximizing for domain invariance.
\newblock 2014.

\bibitem{cite:ICML15DAN}
M.~Long, Y.~Cao, J.~Wang, and M.~I. Jordan.
\newblock Learning transferable features with deep adaptation networks.
\newblock In {\em ICML}, 2015.

\bibitem{cite:ICML15RevGrad}
Y.~Ganin and V.~Lempitsky.
\newblock Unsupervised domain adaptation by backpropagation.
\newblock In {\em ICML}, 2015.

\bibitem{cite:ICCV15SDT}
E.~Tzeng, J.~Hoffman, N.~Zhang, K.~Saenko, and T.~Darrell.
\newblock Simultaneous deep transfer across domains and tasks.
\newblock In {\em ICCV}, 2015.

\bibitem{cite:CVPR16DRL}
K.~He, X.~Zhang, S.~Ren, and J.~Sun.
\newblock Deep residual learning for image recognition.
\newblock In {\em CVPR}, 2016.

\bibitem{cite:TNN11TCA}
S.~J. Pan, I.~W. Tsang, J.~T. Kwok, and Q.~Yang.
\newblock Domain adaptation via transfer component analysis.
\newblock {\em TNNLS}, 22(2):199--210, 2011.

\bibitem{cite:TPAMI12DTMKL}
L.~Duan, I.~W. Tsang, and D.~Xu.
\newblock Domain transfer multiple kernel learning.
\newblock {\em TPAMI}, 34(3):465--479, 2012.

\bibitem{cite:ICML13TCS}
K.~Zhang, B.~Sch\"{o}lkopf, K.~Muandet, and Z.~Wang.
\newblock Domain adaptation under target and conditional shift.
\newblock In {\em ICML}, 2013.

\bibitem{cite:NIPS14FTL}
X.~Wang and J.~Schneider.
\newblock Flexible transfer learning under support and model shift.
\newblock In {\em NIPS}, 2014.

\bibitem{cite:ECCV10Office}
K.~Saenko, B.~Kulis, M.~Fritz, and T.~Darrell.
\newblock Adapting visual category models to new domains.
\newblock In {\em ECCV}, 2010.

\bibitem{cite:CVPR12GFK}
B.~Gong, Y.~Shi, F.~Sha, and K.~Grauman.
\newblock Geodesic flow kernel for unsupervised domain adaptation.
\newblock In {\em CVPR}, 2012.

\bibitem{cite:NIPS14LSDA}
J.~Hoffman, S.~Guadarrama, E.~Tzeng, R.~Hu, J.~Donahue, R.~Girshick,
  T.~Darrell, and K.~Saenko.
\newblock {LSDA}: Large scale detection through adaptation.
\newblock In {\em NIPS}, 2014.

\bibitem{cite:JMLR11MTLNLP}
R.~Collobert, J.~Weston, L.~Bottou, M.~Karlen, K.~Kavukcuoglu, and P.~Kuksa.
\newblock Natural language processing (almost) from scratch.
\newblock {\em JMLR}, 12:2493--2537, 2011.

\bibitem{cite:TPAMI13DLSurvey}
Y.~Bengio, A.~Courville, and P.~Vincent.
\newblock Representation learning: A review and new perspectives.
\newblock {\em TPAMI}, 35(8):1798--1828, 2013.

\bibitem{cite:ICML11DADL}
X.~Glorot, A.~Bordes, and Y.~Bengio.
\newblock Domain adaptation for large-scale sentiment classification: A deep
  learning approach.
\newblock In {\em ICML}, 2011.

\bibitem{cite:CVPR13MidLevel}
M.~Oquab, L.~Bottou, I.~Laptev, and J.~Sivic.
\newblock Learning and transferring mid-level image representations using
  convolutional neural networks.
\newblock In {\em CVPR}, June 2013.

\bibitem{cite:ICML11MDL}
J.~Ngiam, A.~Khosla, M.~Kim, J.~Nam, H.~Lee, and A.~Y. Ng.
\newblock Multimodal deep learning.
\newblock In {\em ICML}, 2011.

\bibitem{cite:COLT09DAT}
Y.~Mansour, M.~Mohri, and A.~Rostamizadeh.
\newblock Domain adaptation: Learning bounds and algorithms.
\newblock In {\em COLT}, 2009.

\bibitem{cite:ML10DAT}
S.~Ben-David, J.~Blitzer, K.~Crammer, A.~Kulesza, F.~Pereira, and J.~W.
  Vaughan.
\newblock A theory of learning from different domains.
\newblock {\em MLJ}, 79(1-2):151--175, 2010.

\bibitem{cite:MM07ASVM}
J.~Yang, R.~Yan, and A.~G. Hauptmann.
\newblock Cross-domain video concept detection using adaptive svms.
\newblock In {\em MM}, pages 188--197. ACM, 2007.

\bibitem{cite:ICML09DAM}
Lixin Duan, Ivor~W Tsang, Dong Xu, and Tat-Seng Chua.
\newblock Domain adaptation from multiple sources via auxiliary classifiers.
\newblock In {\em ICML}, pages 289--296. ACM, 2009.

\bibitem{cite:TPAMI12AMKL}
L.~Duan, D.~Xu, I.~W. Tsang, and J.~Luo.
\newblock Visual event recognition in videos by learning from web data.
\newblock {\em TPAMI}, 34(9):1667--1680, 2012.

\bibitem{cite:NIPS12CNN}
A.~Krizhevsky, I.~Sutskever, and G.~E. Hinton.
\newblock Imagenet classification with deep convolutional neural networks.
\newblock In {\em NIPS}, 2012.

\bibitem{cite:JMLR12MMD}
A.~Gretton, K.~Borgwardt, M.~Rasch, B.~Sch{\"o}lkopf, and A.~Smola.
\newblock A kernel two-sample test.
\newblock {\em JMLR}, 13:723--773, March 2012.

\bibitem{cite:NIPS04SSLEM}
Y.~Grandvalet and Y.~Bengio.
\newblock Semi-supervised learning by entropy minimization.
\newblock In {\em NIPS}, 2004.

\bibitem{cite:CVPR15BCNN}
Tsung-Yu Lin, Aruni RoyChowdhury, and Subhransu Maji.
\newblock Bilinear cnn models for fine-grained visual recognition.
\newblock In {\em CVPR}, pages 1449--1457, 2015.

\bibitem{cite:AAAI16CORAL}
B.~Sun, J.~Feng, and K.~Saenko.
\newblock Return of frustratingly easy domain adaptation.
\newblock In {\em AAAI}, 2016.

\bibitem{cite:CVPR11DB}
A.~Torralba and A.~A. Efros.
\newblock Unbiased look at dataset bias.
\newblock In {\em CVPR}, 2011.

\end{thebibliography}
\bibliographystyle{unsrt}
\end{small}

\end{document}